\documentclass{article}

\usepackage{etoolbox}
\newbool{isArxiv}
\setbool{isArxiv}{true} 
    \PassOptionsToPackage{numbers, compress}{natbib}
\ifbool{isArxiv}{
    \usepackage[preprint]{neurips_2024}
}{
    \usepackage{neurips_2024}
}

\newif\ifreview 
\newif\ifarxiv 
\newif\ifcamera 
\newif\ifrebuttal

\newcommand{\mypara}[1]{\vspace{1mm}\noindent\textbf{#1}~~}

\newcommand{\ky}[1]{{#1}}

\usepackage[utf8]{inputenc} % allow utf-8 input
\usepackage[T1]{fontenc}    % use 8-bit T1 fonts
\usepackage{hyperref}       % hyperlinks
\usepackage{url}            % simple URL typesetting
\usepackage{booktabs}       % professional-quality tables
\usepackage{amsfonts}       % blackboard math symbols
\usepackage{nicefrac}       % compact symbols for 1/2, etc.
\usepackage{microtype}      % microtypography
\usepackage{xcolor}         % colors

\usepackage{multirow}
\usepackage{adjustbox}
\usepackage{amsmath,amssymb}
\usepackage{multirow}
\usepackage{tikz}
\usepackage{subcaption}
\usepackage[font=small]{caption}
\usepackage[super]{nth}
\usepackage{algorithm}
\usepackage{algpseudocode}
\usepackage{tabularx}
\usepackage{makecell}
\usepackage{enumitem}

\usepackage{bbding}
\usepackage{threeparttable}
\usepackage{wrapfig}
\usepackage{colortbl}
\usepackage{graphicx}
\graphicspath{ {./figures/}, {./figures/image_experiment/}, {./figures/tex/} }

\title{Unleashing Generalization of End-to-End Autonomous Driving with Controllable Long Video Generation}

\author{
    Enhui Ma$^{1,2,3}$\thanks{Co-first authors}\enskip\enskip 
    Lijun Zhou$^{2}$\footnotemark[1]\enskip\enskip
    Tao Tang$^{4,2}$\enskip\enskip
    Zhan Zhang$^{5,1}$\enskip\enskip
    Dong Han$^{6,1}$\enskip\enskip
    Junpeng Jiang$^{7,2}$\enskip\enskip \\
    {\bfseries Kun Zhan$^{2}$}\enskip\enskip
    {\bfseries Peng Jia$^{2}$}\enskip\enskip
    {\bfseries Xianpeng Lang$^{2}$}\enskip\enskip
    {\bfseries Haiyang Sun$^{2}$}\enskip\enskip
    {\bfseries Di Lin$^{3}$} \enskip\enskip
    {\bfseries Kaicheng Yu$^{1}$}\thanks{Corresponding Author}\\
    \textsuperscript{1}Westlake University \enskip 
    \textsuperscript{2}Li Auto Inc. \enskip
    \textsuperscript{3}Tianjin University \enskip
    \textsuperscript{4}Shenzhen Campus, Sun Yat-sen University \enskip \\
    \textsuperscript{5}Southeast University \enskip 
    \textsuperscript{6}Harbin Engineering University \enskip
    \textsuperscript{7}Harbin Institute of Technology(Shenzhen) \enskip \\
    \texttt{\{maenhui, kyu\}@westlake.edu.cn, zhoulijun@lixiang.com}
}
\begin{document}

\maketitle

\begin{figure*}[h!]
\centering
\vspace{-0.9cm}
\includegraphics[width=\linewidth]{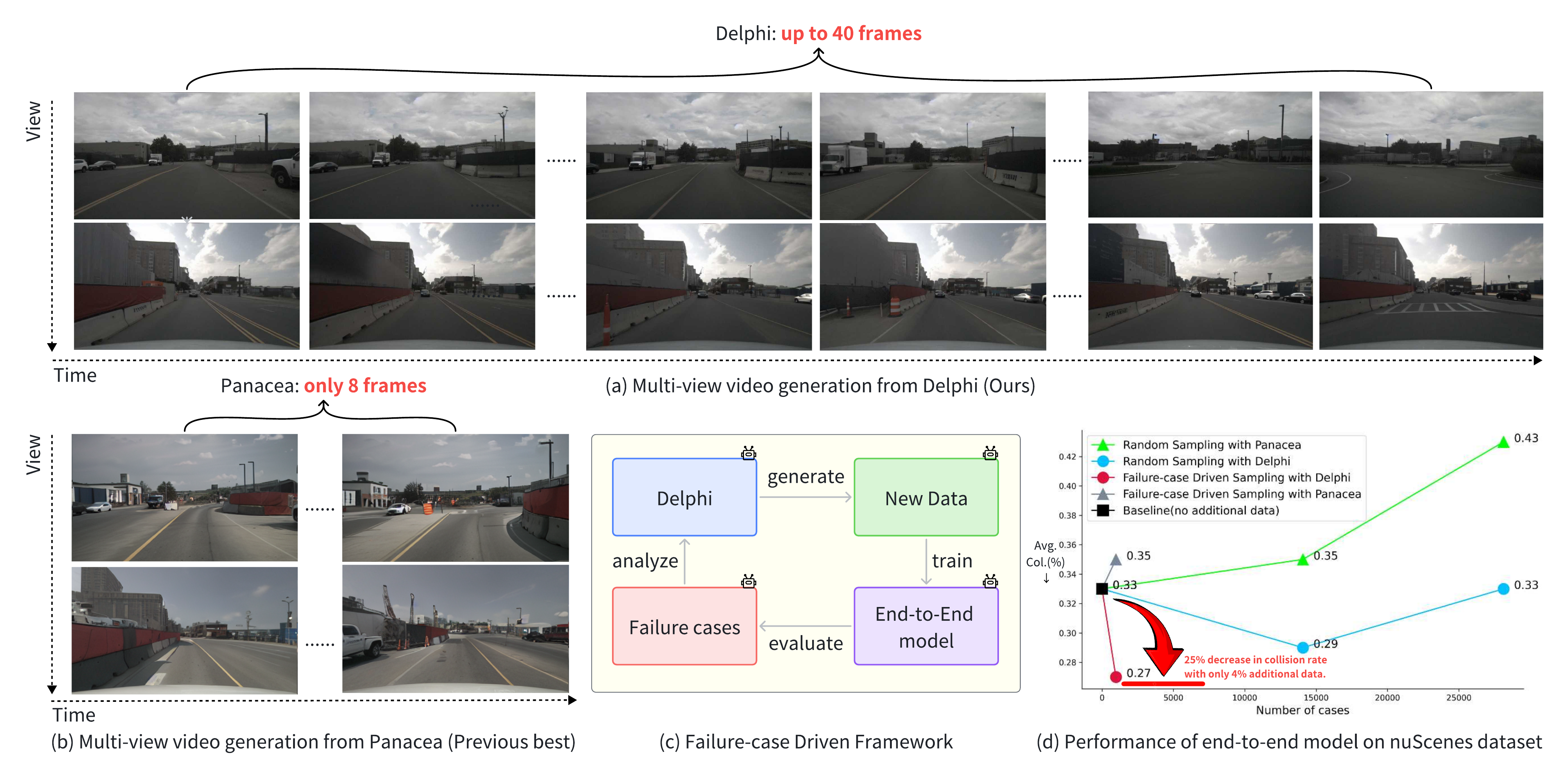}
\vspace{-0.5cm}
\caption{\textbf{Overview of our method.} We show that \textbf{(a)} our \emph{Delphi} can generate up to 40 frames consecutive videos while \textbf{(b)} existing best only generate 8 frames. \textbf{(c)} With the failure-cased driven framework equipped with \emph{Delphi}, \textbf{(d)} we can significantly boost the end-to-end model performance with much smaller cost. }
\label{fig:teaser}
\vspace{-0.3cm}
\end{figure*}

\begin{abstract}

% Generative models have recently garnered increasing attention due to their impressive ability to produce realistic and diverse data. There exist some methods exploring leveraging generative models to synthesize training data to enhance perception models.

% As collecting long-tail scenarios in autonomous driving, such as potential collision, is both dangerous and unethical, developing generative models to synthesize realistic data has drawn increasing attention recently. 

% however, whether these data can be used to increase the p} 
% However, end-to-end autonomous driving models still face significant challenges in discerning the scale and patterns of data that are truly critical for effective decision-making. 
% To this end, we propose \emph{Delphi}, a novel controllable long video generation method with spatiotemporal consistency 
Using generative models to synthesize new data has become a de-facto standard in autonomous driving to address the data scarcity issue. Though existing approaches are able to boost perception models, we discover that these approaches fail to improve the performance of planning of end-to-end autonomous driving models as the generated videos are usually less than 8 frames and the spatial and temporal inconsistencies are not negligible. To this end, we propose \emph{Delphi}, a novel diffusion-based long video generation method with a shared noise modeling mechanism across the multi-views to increase spatial consistency, and a feature-aligned module to achieves both precise controllability and temporal consistency. Our method can generate up to 40 frames of video without loss of consistency which is about 5 times longer compared with state-of-the-art methods. Instead of randomly generating new data, we further design a sampling policy to let \emph{Delphi} generate new data that are similar to those failure cases to improve the sample efficiency. This is achieved by building a failure-case driven framework with the help of pre-trained visual language models. Our extensive experiment demonstrates that our \emph{Delphi} generates a higher quality of long videos surpassing previous state-of-the-art methods. Consequentially, with only generating 4\% of the training dataset size, our framework is able to go beyond perception and prediction tasks, for the first time to the best of our knowledge, boost the planning performance of the end-to-end autonomous driving model by a margin of 25\%. 
Please see visual demos at \url{https://westlake-autolab.github.io/delphi.github.io/}.

% https://anonymous-github-8ab1cv.github.io/. 
% Specifically, \emph{Delphi} maintains long-range consistency of videos by integrating a novel Noise Reinitialization and Feature-aligned Temporal Consistency module, and provides fine-grained control at both scene-level and object-level by injecting rich control signals using cross attention.
% Noticing that existing methods usually uniformly random generating videos which is potentially inefficient, 

% automatically identify data patterns involved in these bad cases and retrieves relevant data from only the training dataset with the same patterns for generating new data to train end-to-end models.
% Then to tackle the current dilemma of end-to-end models, we propose a failure-case driven framework based on \emph{Delphi}, which primarily discerns the scale and patterns of critical data from failure cases.
% The framework 

% only with a small amount of diversity critical data generated from \emph{Delphi}, our failure-case driven framework significantly improves the generalization performance of end-to-end models. 

\end{abstract}
\section{Introduction}

End-to-end autonomous driving has recently garnered increasing attention~\cite{hu2023uniad, Jiang2023VADVS, Yang2023VisualPC}, which directly learns to plan motions from raw sensor data, reducing heavy reliance on hand-crafted rules and avoiding cascading modules. 
However, current end-to-end models face significant challenges in the scale and quality of training data. 
Insufficient data diversity can lead to model overfitting~\cite{zhai2023rethinking}, such as the collected real-word trajectories mainly involve straight lines for the "go straight" action, when applied to more complex scenarios like "turn left on cross intersection", the model is prone to fail.
While large-scale and high-quality annotated data is crucial for the safe and robust end-to-end autonomous driving system, unfortunately, collecting such data poses challenges, particularly in situations involving dangerous scenes where data collection can be difficult or unsafe.

Although recent generative models~\cite{gao2023magicdrive,wen2023panacea,yang2023bevcontrol} have gained remarkable progress in mitigating the problem of data scarcity for perception models, which is achieved by employing ControlNet~\cite{zhang2023_controlnet} to control the geometric position of scene elements with injected BEV layouts and extend across the view dimension to generate multi-view images.
% Recent advancements in generative models~\cite{wen2023panacea, gao2023magicdrive, wang2023drivedreamer, yang2023bevcontrol} have shown promising data generation abilities and significantly enhanced various downstream perception tasks, such as 3D detection~\cite{li2022_bevformer,liu2023_bevfusion,wang2023streampetr}, map segmentation~\cite{zhou2022_CVT}, and multi-object tracking~\cite{wang2023streampetr}. 
% Although recent generative models~\cite{gao2023magicdrive,wen2023panacea,yang2023bevcontrol} have gained remarkable progress in mitigating the problem of data scarcity for perception models, which is achieved by employing ControlNet~\cite{zhang2023_controlnet} to control the geometric position of scene elements with injected BEV layouts and extend across the view dimension to generate multi-view images.
When applied to end-to-end autonomous driving which requires long multi-view videos, two main challenges arise: \textbf{spatial-temporal consistency} and \textbf{precise controllability}. 
Existing generative methods~\cite{gao2023magicdrive,wen2023panacea,wang2023drivedreamer,wang2023driveWM} simply utilize cross-frame attention with the previously generated frame to ensure consistency, which overlooks the differences in noise patterns between image generation and video generation, as well as the alignment of features in the cross-frame attention. 
As a result, temporal consistency can only be maintained in short video sequences, such as Panacea~\cite{wen2023panacea} with 8 frames and MagicDrive~\cite{gao2023magicdrive} with 7 frames. 
Furthermore, current methods only offer coarse-grained control over the generated videos, limited to modifying simple global attributes, such as changing weather with simple text prompts. They cannot finely control the overall architectural style of the scene or the specific appearance attributes of individual objects. 

% \ky{We notice the existing approaches are adding noises in an independent manner for different views, and does not consider the cross view consistencies. }
To this end, we propose a novel multi-view long video generation method, dubbed \emph{Delphi}, to address these limitations. First, we notice that existing approaches fall short in two aspects: i) adding independent noise to different views and does not consider the cross-view consistency; ii) exploiting a simple cross attention to fuse multiple feature spaces with different reception fields. We then propose two components, a noise reinitialization module to model the shared noise across frames and a feature-aligned temporal consistent module to address the second challenge. 
% In this paper, we propose an innovative multi-view long video generation method, \emph{Delphi}, which tackles aforementioned issues with Noise Reinitialization Module (NRM) and Feature-aligned Temporal Consistency Module (FTCM), as illustrated in Figure~\ref{fig:delphi-framework}. 
% Building on the original independent noise, 
% Specifically, considering the natural temporal correlation of videos, NRM applies the same base noise across both spatial and temporal dimensions respectively, ultimately obtaining initial noise that is spatiotemporally correlated. Then, FTCM ensures global consistency between consecutive frames by performing attention interactions with the latent features at the same network depth from the previous frame. Besides, FTCM uses object-level cross-attention with object masks from consecutive frames to maintain the consistency of objects. Additionally, \emph{Delphi} injects the dense text prompts obtained from pre-trained VLMs for the scene and individual objects by cross attention, which allows users to control scene content from a coarse to fine-grained level, thereby enhancing the diversity of the generated videos.

Leveraging \emph{Delphi} as the data engine, we further propose a failure-case driven framework, which automatically enhances the generalization of end-to-end models.
Specifically, this framework integrates several steps as shown in Figure~\ref{fig:teaser}: 1) evaluating the end-to-end model, collecting failure cases, 2) analyzing the implicit data patterns using pre-trained VLMs, 3) retrieving similar patterned data from existing training data, 4) generating diverse training data with \emph{Delphi}, and then updating the end-to-end model.
To investigate the effectiveness of our method, we conducted extensive experiments on the large-scale public dataset nuScenes~\cite{caesar2020_nuscenes}. 
Firstly, comprehensive evaluations with various metrics demonstrate that our \emph{Delphi} generate 
high-quality long multi-view videos with spatiotemporal consistency and precise controllability.
Furthermore, the proposed failure-case driven framework achieves remarkable improvement in the generalization capability of end-to-end models at a low cost. 
% Specifically, iterating the failure-case driven framework significantly reduces the collision rate of the popular end-to-end model, UniAD, by 25 \%.

Our contributions can be summarized as follows: 
\begin{itemize}[nolistsep, leftmargin=0.5cm]
\setlength{\itemsep}{0.05cm}
\setlength{\parsep}{0pt}
\setlength{\parskip}{0pt}

\item We introduce \emph{Delphi}, a novel method that can generate up to 12 seconds~(40 frames) temporally consistent multi-view videos in autonomous driving~(AD) scenarios, which is 5x longer compared to the state-of-the-art video generation methods. In addition, \emph{Delphi} encompasses the control ability including both object and scene level details to enrich the diversity of generated data.
% We introduce \emph{Delphi}, a innovative controllable long video generation method, which generates temporally consistent long-term multi-view videos (\ky{up to} 12 seconds or 40 frames) for autonomous driving. 
% To the best of our knowledge, \emph{Delphi} produces the longest videos among existing autonomous driving video generative models. Additionally, \emph{Delphi} offers fine-grained control at both the object-level and scene-level, which enhances the diversity of the training data.
% , leading to more robust and comprehensive models.

% \item 
% We propose a failure-case driven framework with \emph{Delphi} as the data engine, which automatically analyzes failure cases, retrieves training data, generates diverse training data based on our  \emph{Delphi}, and finetunes to improve the end-to-end models.
% It can retrieve data from existing training sets that match the patterns of failed cases, driving the generation model to enrich the diversity of training data. This allows for fine-tuning of the end-to-end model.
\item We propose a failure-case driven framework to drastically increase sampling efficiency. We show that using a long-term video generation method that trained purely on the training dataset, we are able to improve the UniAD's performance by 25\% (collision rate reduces from 0.34 to 0.27) from with generating only 4\% (972 cases) of the training dataset size. 
\item Compared to earlier works that only manage to improve the perception ability using synthetic data, we are the first, to the best of our knowledge, to showcase that a data engine can go beyond the perception task and automatically improve the planning ability of end-to-end AD methods. We hope this can shed light on alleviating the long-tail issue of the large-scale development of AD vehicles. 

% Extensive experiments demonstrate that our \emph{Delphi} generate high-quality long multi-view videos, and the failure-case driven framework with \emph{Delphi} significantly improves the performance of end-to-end models with a low cost.

\end{itemize}
\section{Related work}
\label{related}

\mypara{End-to-end Autonomous Driving.}
End-to-end models have garnered significant attention in the field of autonomous driving. These models simplify the traditional modular pipeline by integrating perception, prediction, decision, and planning into a single learning framework. TransFuser\cite{prakash2021transfuser} fuses visual and lidar inputs with a transformer-based architecture to improve perception and driving decisions. ST-P3\cite{hu2022stp3} leverages spatial-temporal feature learning to improve perception, prediction, and planning tasks. UniAD\cite{hu2023uniad} effectively combines multiple perception and prediction tasks to improve planning performance.  VAD \cite{Jiang2023VADVS} explores the potential of vectorized scene representation for planning and getting rid of dense maps. VADv2\cite{chen2024vadv2} utilizes probabilistic planning to manage uncertainties and transforms multi-view image sequences into environmental token embeddings to predict and sample vehicle control actions. In this paper, we have opted to utilize the well-known UniAD as our downstream model due to the computational resource constraints. 

% \subsection{Video Generation}
% 和World Models融为一体
% \eh{zlj}
\mypara{Generative model to boost autonomous driving performance.} Video generation stands as a pivotal technology in understanding the visual world. Early methods mainly include Variational Autoencoders (VAEs) \cite{denton2018stochastic,hyvarinen2005estimation}, flow-based models \cite{kumar2019videoflow}, and Generative Adversarial Networks (GANs) \cite{mathieu2015deep, saito2017temporal, tulyakov2018mocogan, vondrick2016generating}.
Notably, the recent achievements of diffusion models in image generation \cite{nichol2021_glide, rombach2022_ldm, ruiz2023dreambooth} have stimulated interest in their application to video generation \cite{harvey2022flexible, hoppe2022diffusion}. Diffusion-based methods have significantly improved realism, controllability, and temporal consistency. With their controllable attributes, text-based conditional video generation has garnered increasing attention, leading to the emergence of numerous methods \cite{rombach2022_ldm, ho2022imagen, singer2022make, wu2022_Tune_A_Video, zhou2022magicvideo}.
Especially popular models like diffusion-based models~\cite{rombach2022_ldm, song2020_ddim, ho2020_ddpm, zhang2023_controlnet}, which enable users to generate images with controllability.
Inspired by this innovation, some models~\cite{gao2023magicdrive,yang2023bevcontrol} employ ControlNet to control the geometric position of scene elements by injecting BEV layouts and extend this approach across the view dimension to generate multi-view images.
% , ensuring consistency between views with the addition of cross-view attention. 
The other models~\cite{wen2023panacea,wang2023drivedreamer,zhao2024drivedreamer-2} further extend this to the temporal dimension to generate multi-view videos,
% maintaining consistency between frames with the cross-frame attention, 
which are all trained based on the pre-trained image models~\cite{rombach2022_ldm}. 
BEVGen \cite{swerdlow2023_bevgen} specializes in generating multi-view street images based on Bird’s Eye View (BEV) layouts. BEVControl \cite{yang2023bevcontrol} proposes a two-stage generation pipeline for creating image foregrounds and backgrounds from BEV layouts. DriveDreamer \cite{wang2023drivedreamer} and Panacea \cite{wen2023panacea} introduce a layout-conditioned video generation system aimed at diversifying data sources for training perception models. GAIA-1 \cite{hu2023gaia} and ADriver-I \cite{jia2023adriver} integrate large language models into video generation; a concurrent work DriveDreamer-2~\cite{zhao2024drivedreamer-2} proposes a traffic simulation pipeline employing only text prompts as input, which can be utilized to generate diverse traffic conditions for driving video generation, however, it requires one frame input and does not work in the setting of boosting planning task. All in all, these methods can only generate fairly short videos up to 8 frames, while our \emph{Delphi} can generate much longer ones.
% The text must be confined within a rectangle 5.5~inches (33~picas) wide and
% 9~inches (54~picas) long. The left margin is 1.5~inch (9~picas).  Use 10~point
% type with a vertical spacing (leading) of 11~points.  Times New Roman is the
% preferred typeface throughout, and will be selected for you by default.
% Paragraphs are separated by \nicefrac{1}{2}~line space (5.5 points), with no
% indentation.

% The paper title should be 17~point, initial caps/lower case, bold, centered
% between two horizontal rules. The top rule should be 4~points thick and the
% bottom rule should be 1~point thick. Allow \nicefrac{1}{4}~inch space above and
% below the title to rules. All pages should start at 1~inch (6~picas) from the
% top of the page.

% For the final version, authors' names are set in boldface, and each name is
% centered above the corresponding address. The lead author's name is to be listed
% first (left-most), and the co-authors' names (if different address) are set to
% follow. If there is only one co-author, list both author and co-author side by
% side.

% Please pay special attention to the instructions in Section \ref{others}
% regarding figures, tables, acknowledgments, and references.

\section{Method}
\label{method}

In this section, we first present \emph{Delphi}, an innovative method for generating long multi-view videos of autonomous driving. 
There are two core modules designed for generating temporally consistent videos: Noise Reinitialization Module (NRM) in Sec ~\ref{sec:Noise Reinitialization} and Feature-aligned Temporal Consistency Module (FTCM) in Sec ~\ref{sec:Feature-aligned Temporal Consistency}.
Finally, Sec.~\ref{sec:Failure-case Driven Framework} presents a failure-case driven framework to show how we can leverage the long-term video generation ability to automatically enhance the generalization of an end-to-end model with only data from the training dataset.

\begin{figure}[t]
    \centering
    \vspace{-0.9cm}
      \includegraphics[width=1.0\linewidth]{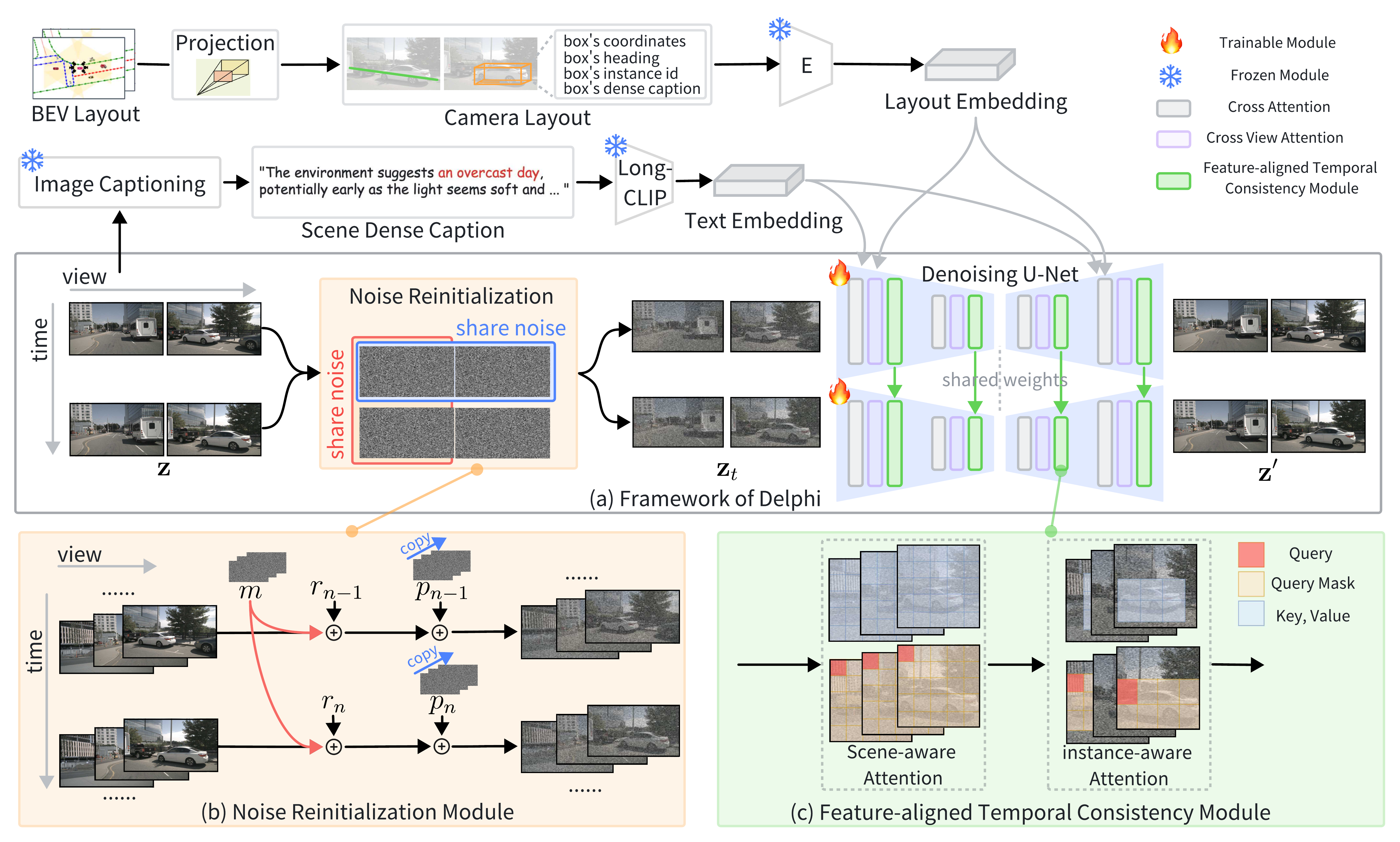}
    \vspace{-0.5cm}
    \caption{\textbf{(a) Architecture of \emph{Delphi}.} It takes multi-view videos $\mathbf{z}$ and the corresponding BEV (Bird's Eye View) layout sequences as input. Each video consists of \( N \) frames and \( V \) views. The BEV layout sequences are first projected into camera space according to camera parameters, resulting in camera layouts that include both foreground and background layouts. 
Specifically, the foreground layout includes the bounding box's corner coordinates, heading, instance id, and dense caption, while the background one includes different colored lines to represent road trends. 
The layout embeddings, processed by the encoder, are injected into the U-Net through cross-attention to achieve fine-grained layout control in the generation process. Additionally, we leverage VLM~\cite{achiam2023gpt} to extract dense captions for the input scenes,  which are encoded by Long-CLIP~\cite{zhang2024longclip} to obtain text embeddings, which are then injected into the U-Net via text cross-attention to achieve text-based control. 
We further design two key modules,  
\textbf{(b) Noise Reinitialization Module} that encompass a share noise across different views  and \textbf{(c) Feature-aligned Temporal Consistency Module} to ensure spatial and temporal consistency accordingly.
}
    \label{fig:delphi-framework} 
    \vspace{-0.5cm}
\end{figure}

\subsection{Delphi: A Controllable Long Video Generation Method}

Here, we present the architecture of \emph{Delphi} in Figure~\ref{fig:delphi-framework}(a).  Existing models tend to overlook the noise formulation across time and spatial dimensions, leading to inferior long-video generation quality. In contrast, we propose two key components to address these challenges: a noise reinitialization module and a feature-aligned temporal consistency module. 

\subsubsection{Noise Reinitialization Module}
\label{sec:Noise Reinitialization}
Multi-view videos naturally exhibit similarities across both time and view dimensions. However, existing approaches are categorized into two groups, i) concurrent single-view video generation methods~\cite{luo2023videofusion,ren2024consisti2v,qiu2023freenoise} cannot be directly applied in outdoor multi-view scenarios; ii) multi-view generative models are adding independent noise that does not consider cross view consistencies~\cite{gao2023magicdrive,yang2023bevcontrol,wen2023panacea}. Here, we plan to address this issue by introducing a shared noise across these two dimensions.
% Thus, we propose a more effective method to introduce shared noise along these two dimensions. 
% Thus, we propose a more effective method to introduce shared noise along these two dimensions, drawing inspiration from recent advancements in video generation~\cite{luo2023videofusion,ren2024consisti2v,qiu2023freenoise}.
Specifically, 
% instead of adding independent noise as in previous methods, 
as shown in Figure~\ref{fig:delphi-framework}(b), we introduce shared motion noise $m$ along the temporal dimension and shared panoramic noise $p$ along the viewpoint dimension. 
This results in a noisy version of the multi-view video that is correlated across both time and view dimensions.
The process of incorporating shared noise can be represented as follows:

	\begin{equation}
		\mathbf{z}_n^v = \sqrt{\hat{\alpha}}x_n^v + \sqrt{1 - \hat{\alpha}}(r_n^v + m^v + p_n),
	\end{equation}

where $\mathbf{z}_n^v \in \mathbb{R}^{1 \times 1 \times h \times w}$ represents the image latent variable of view $v$ at frame $n$, $m \in \mathbb{R}^{V \times 1 \times h \times w}$ are the shared motion noise of $V$ views, and $p \in \mathbb{R}^{1 \times N \times h \times w}$ are the shared panoramic noise of $N$ frames. For simplicity, we omit the subscript \( t \).

\begin{figure}[t]
    \centering
    \vspace{-0.7cm}
      \includegraphics[width=1.0\linewidth]{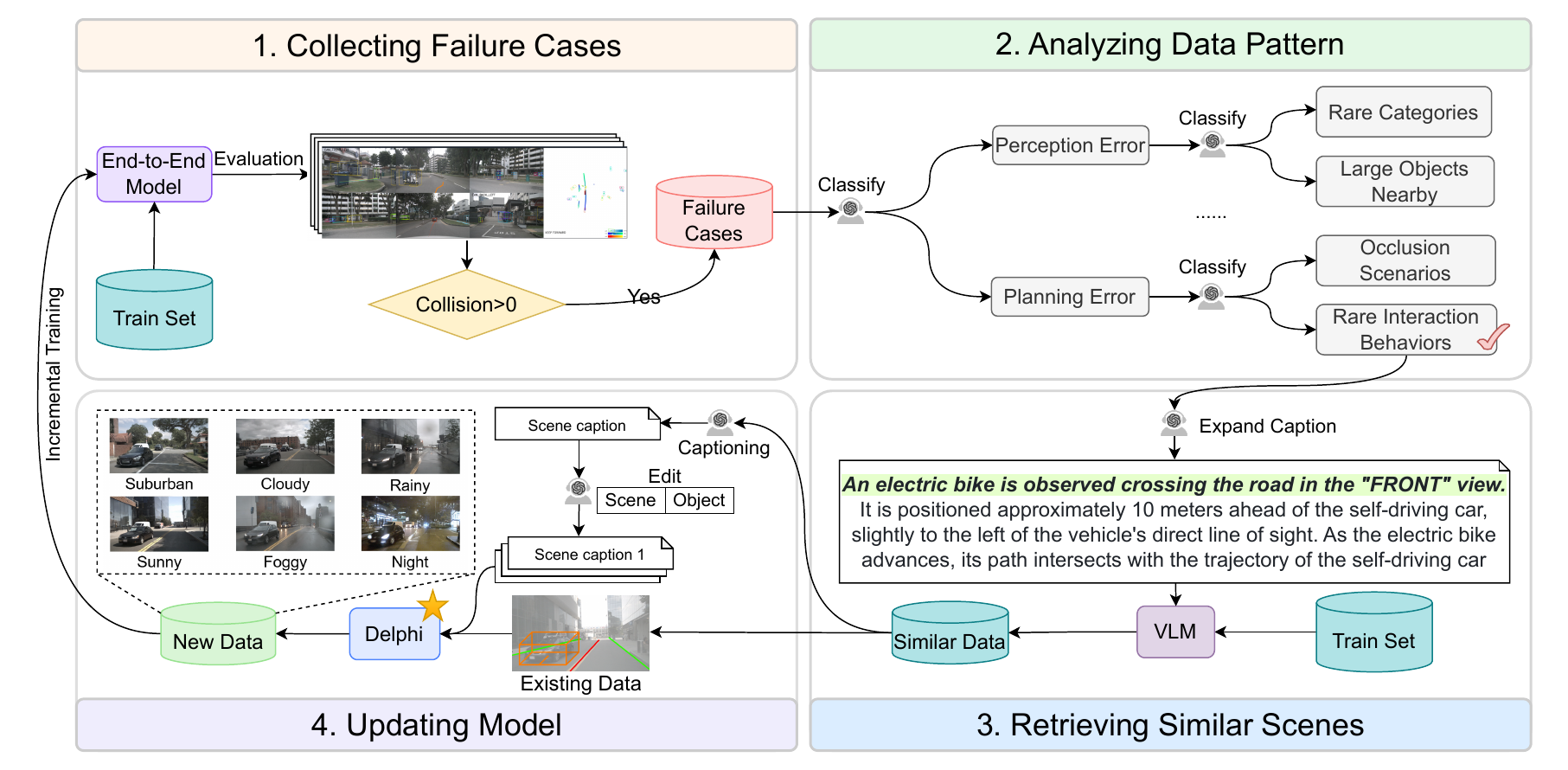}
    \vspace{-0.5cm}
    \caption{\textbf{Overview of Failure-case Driven Framework.}    }
    \label{fig:failure-case-driven-framework} 
    \vspace{-0.2cm}
\end{figure}

\subsubsection{Feature-aligned Temporal Consistency}
\label{sec:Feature-aligned Temporal Consistency}

% The feature-aligned temporal consistency-based U-Net structure is specifically designed to generate temporally coherent long videos. Achieving this goal typically involves using cross-frame attention to generate multi-view videos based on diffusion models. 
Existing methods~\cite{wen2023panacea,gao2023magicdrive,wang2023driveWM}, when generating the current frame, exploits a simple cross-attention mechanism to fuse previous frame information into the current view. However, they tend to overlook the fact that 
% is usually established with the final output of the previous frame to transfer the relevant features. 
% However, they neglect an essential aspect: 
features located at different network depths possess varying receptive fields. Consequently, this coarse feature interaction method fails to capture all the information from the receptive fields from different levels of the previous frame, leading to sub-optimal video generation performance.

To this end, we propose a more effective structure to fully establish feature interactions between aligned features at the same network depth in adjacent frames as shown in Figure~\ref{fig:delphi-framework}~(c).
We approach this by ensuring global consistency and optimizing local consistency, incorporating two main designs: Scene-aware Attention and Instance-aware Attention.

\paragraph{Scene-aware Attention.}
To fully leverage the rich information from features at different network depths in the previous frame, we propose a scene-level cross-frame attention mechanism. Specifically, this module performs the attention calculation on features at the same network depth between adjacent frames. The computation process can be represented as follows:
\begin{equation}
     \text{Attn}_{\text{scene}}(Q_n^i, K_{n-1}^i, V_{n-1}^i) = \text{softmax}\left(\frac{{Q_n^i}{K_{n-1}^i}^T}{\sqrt{d_k}}\right)V_{n-1}^i,
\end{equation} 
where \( Q_n^i \) (query) is the latent feature map from the current frame $n$ at a specific network depth $i$, \( K_{n-1}^i \) (key) and \( V_{n-1}^i \) (value) are the latent feature maps from the previous frame $n-1$ at the same network depth $i$, and \( d_k \) is the dimensionality of the key. And $i = {1, ..., I}$, where $I$ indicates the total number of U-Net blocks.   We omit the view channel for simplicity.
By applying this scene-aware attention mechanism, the module effectively transfers the global style information from the previous frame to the current frame, ensuring temporal consistency across frames.

\paragraph{Instance-aware Attention.}
To enhance the coherence of moving objects within the scene, we propose an Instance-aware Cross-frame Attention mechanism. Compared to scene-level attention, this module uses foreground bounding boxes as attention masks to compute feature interactions in local regions between adjacent frames. The computation process can be represented as follows:

\begin{align}
    \text{Attn}_{\text{ins}}(Q_n^i, K_{n-1}^i, V_{n-1}^i) &= \text{softmax}\left(\frac{({Q_n^i \cdot M_n}) ({{K_{n-1}^i} \cdot M_{n-1}})^T}{\sqrt{d_k}}\right)({ V_{n-1}^i \cdot M_{n-1}}), \\[3pt]
    \hat{Q}_n^i = Q_n^i &+ \text{Zero}[\text{Attn}_{\text{ins}}(Q_n^i, K_{n-1}^i, V_{n-1}^i)]q,
\end{align} 

where \( M_n \) and  \( M_{n-1} \) are the masks of foreground objects from the current frame and previous frame respectively, focusing on the region defined by the foreground bounding box, and \text{Zero} indicates the trainable convolution layers initialized with a value of 0.

% \paragraph{Spatial Consisitency}
% 1. cross view attention
% 2. instance id and densc caption control

\subsection{Failure-case Driven Framework}
\label{sec:Failure-case Driven Framework}

In order to leverage the generated data, common approaches will randomly sample a subset of the training dataset and then apply video generation models to augment these data to enhance the performance of downstream tasks. We hypothesize that this random sample does not consider the existing distribution of long-tail cases and is substantial for further optimization. We hence propose a simple but effective failure-case driven framework that exploits four steps to reduce the computational costs. As shown in Figure~\ref{fig:failure-case-driven-framework}, we first evaluate the existing failure cases as a starting point, we then implement a visual language-based method to analyze the patterns of these data and retrieve similar scenes to gain a deeper understanding of the context, we then diversify the captions for scene and instance editing, to generate new data with different appearances. Finally, we train the downstream tasks with such additional data for a few epochs to increase the generalization ability.

Note that, all of these operations are conducted on training set to avoid any potential leak of the validation information. Please see supplementary materials for detailed implementation of each component. In addition, we notice a concurrent work~\cite{su2024aide} that exploits a similar idea. However, their approach only works for 2D detection tasks while our method is capable of improving end-to-end planning ability.
% Inspired by AIDE~\cite{su2024aide}, 
% this section introduces our failure-case driven framework, which is rooted in the \emph{Delphi} methodology. We have devised a four-step framework tailored for optimizing end-to-end algorithms. This comprehensive approach involves: 
% i) Evaluating the model's performance, (2) Analyzing data patterns, (3) Retrieving similar scenes to gain a deeper understanding of the context, and (4) Updating the model based on the data gained from the previous steps. This framework is visually represented in Figure~\ref{fig:failure-case driven-framework}.

\section{Experiments}
\label{experiments}
% \vspace{-0.5em} 
% \subsection{Datasets and Metrics}
% \eh{yuxin}
% \vspace{-0.5em} 
% \subsection{Experiment Details}
% We report the dataset and hyperparameters that we used in our experiments. 
% \mypara{Dataset and metrics.} 
We follow popular methods~\cite{wen2023panacea,yang2023bevcontrol}, to use nuScenes~\cite{caesar2020_nuscenes} and use FID~\cite{heusel2017_fid}, FVD~\cite{unterthiner2018fvd}, and downstream model's performances on newly generated data to evaluate the image, video, and sim-to-real gap. See the Appendix for more details.
% See detailed definition in supplementary materials of each metric.

\vspace{-0.2cm}
\paragraph{Dataset.}
We conduct extensive experiments on the popular nuScenes~\cite{caesar2020_nuscenes} validation dataset, which comprises 150 driving scenes marked by dense traffic and intricate street driving scenarios. Each scene contains roughly 40 frames. 
% Key frames are sampled and annotated at 2Hz, resulting in 40 frames per scene. These samples consist of imagery captured by six RGB cameras, offering a complete 360° horizontal Field of View (FOV).
% Out of the 1000 scenes, 700 are allocated for training, 150 for validation, and the remaining for testing.
% Our experiments primarily focus on the the validation set.
We utilize ten foreground categories (i.e., bus, car, bicycle, truck, trailer, motorcycle)
to create detailed street foreground object layouts. Four background classes
% divider, lane boundary, centerline, and pedestrain crossing 
obtained from the map expansion pack are used to generate background layouts.

\mypara{Hyperparameters. } We train our models on 8 A800 80GB GPUs. The diffusion U-Net is optimized using the AdamW~\cite{kingma2014adam} optimizer with a learning rate of 5e-5.
We resize the original images from 1600 × 900 to 512 × 512. During training, the video length is set to 10, and we generate video frames sequentially in a streaming manner. For inference, we use the PLMS~\cite{liu2022pseudo} sampler configured with 50 sampling steps. The spatial resolution of the video samples is set to 512 × 512, with a frame length of 40. The inference length is not restricted and could be 40 or longer. Our model is trained on the nuScenes dataset with 50,000 steps for the cross-view model and 20,000 steps for the temporal model.

\subsection{Comparing Delphi to state-of-the-art video generation methods}

\setlength{\tabcolsep}{9pt}
\renewcommand{\arraystretch}{1.2}
\begin{table*}
\centering
\vspace{-0.1cm}
\caption{We compare Delphi with state-of-the-art methods on the nuScenes validation set. The results measure the spatial-temporal consistency and  controllability of different methods. $\downarrow$/$\uparrow$ means a smaller/larger value of the metric represents a better performance. 
% $*$ means the results computed using the official github release code or rendered videos on validation set. 
}
\vspace{-0.1cm}
\scriptsize
\begin{threeparttable}
\begin{tabular}{l|cc|ccc|cc}
\toprule
\multirow{3}{*}{\textbf{Method}} & \multicolumn{2}{c|}{\textbf{Spatial Consistency}}      & \multicolumn{3}{c|}{\textbf{Temporal Consistency}} & \multicolumn{2}{c}{\textbf{Sim-Real Gap}}              \\ 
\cline{2-8}
                        & \multirow{2}{*}{\textbf{FID$\downarrow$}} & \multirow{2}{*}{\textbf{CLIP$\uparrow$}} & \multicolumn{3}{c|}{\textbf{FVD$\downarrow$}}                  & \multirow{2}{*}{\textbf{NDS$\uparrow$}} & \multirow{2}{*}{\textbf{Avg. Col. Rate$\downarrow$}} \\
                        \cline{4-6}
                        &                      &                       & 4 frames          &  8 frames          & 40 frames          &                      &                           \\ \midrule
BEVGen~\cite{swerdlow2023_bevgen}                  &25.54                     &71.23                       &N/A             &N/A             &Fail              &N/A                      &N/A                           \\
BEVControl~\cite{yang2023bevcontrol}              &24.85                      &82.70                       &N/A             &N/A             &Fail              &N/A                      &N/A                           \\
DriveDreamer~\cite{wang2023drivedreamer}            &26.8                      &N/A                       &N/A             &353.2             &Fail              &N/A                      &N/A                           \\
MagicDrive~\cite{gao2023magicdrive}              &16.20                      &N/A                       &N/A             &N/A             &Fail             &N/A                      &N/A                           \\
MagicDrive*~\cite{gao2023magicdrive}             &46.18                      &82.47                       &617.2             &N/A             &Fail              &34.56                      &3.87                           \\
Panacea~\cite{wen2023panacea}                 &16.96                      &N/A                       &N/A             &139.0             &Fail              &32.10                      &N/A                           \\
Panacea*~\cite{wen2023panacea}                &55.32                      &84.23                       &--             &446.9             &Fail              &29.41                      &1.35                           \\
Drive-WM~\cite{wang2023driveWM}                &15.8                      &N/A                       &N/A             &122.7             &Fail              &N/A                      &N/A                           \\ \rowcolor{gray!15} 
Delphi~(Ours)                  &\textbf{15.08}                      &\textbf{86.73}                       &\textbf{--}             &\textbf{113.5}             &\textbf{275.6}              &\textbf{36.58}                      &\textbf{0.29}                           \\

\bottomrule
\multicolumn{8}{l}{\scriptsize $*$ Results are computed using the official github release code or rendered videos on validation set. }\\
\multicolumn{8}{l}{\scriptsize N/A indicates the model or the pre-trained weights is not open-sourced so we cannot faithfully reproduce.} \\
\end{tabular}
\end{threeparttable}
% \vspace{-0.25in}
\vspace{-0.5cm}
\label{tab:control}
\end{table*}

\begin{figure}[t]
    \centering
    \vspace{-0.7cm}
    \includegraphics[width=1.0\linewidth]{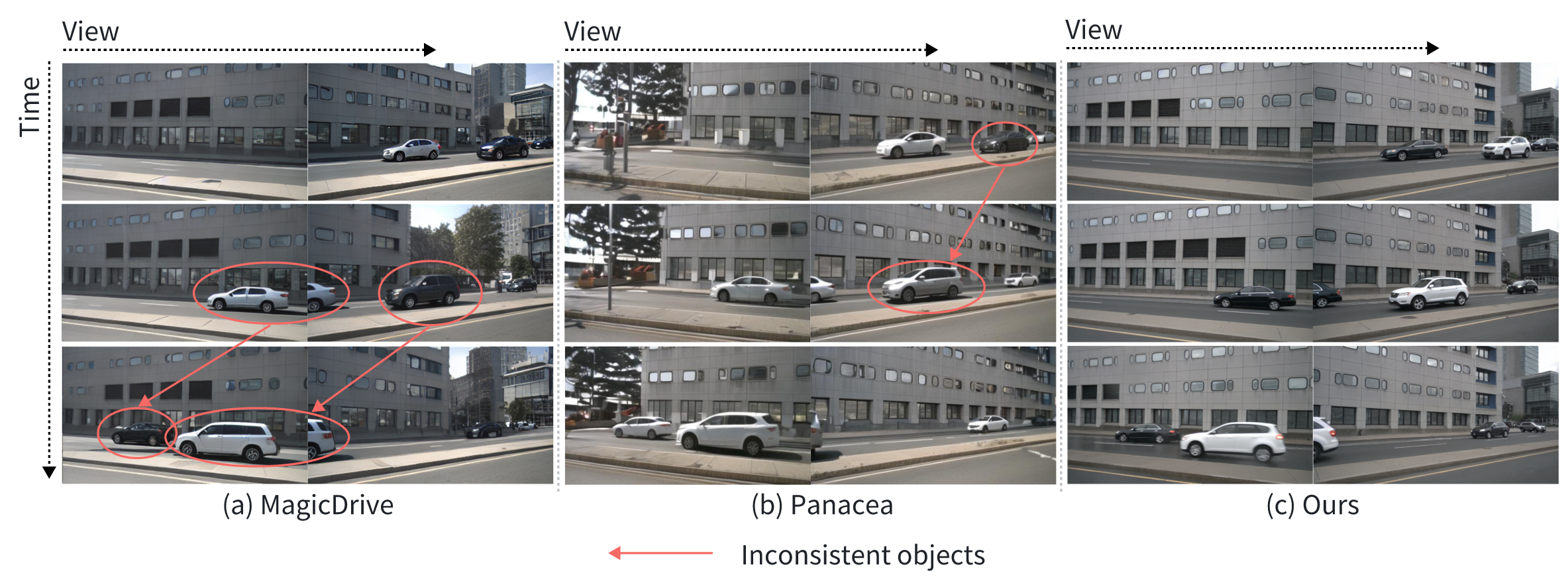}
    \vspace{-0.5cm}
    \caption{\textbf{Visual comparison of local region generated by different generative models.}}
    \vspace{-0.3cm}
    \label{fig:visual conparison of local region} 
    % \vspace{-0.2cm}
\end{figure}

% Quantative~\ref{tab:control} and Qualitive results compared with SOTA methods, such as Panacea, magicdrive......
We assess the quality of video generation through a comprehensive evaluation encompassing both quantitative and qualitative aspects, comparing our approach with previous methodologies.  In Table~\ref{tab:control}, we report the metrics in three aspects on nuScenes validation set, spatial and temporal consistency, and sim-real gap. In short, our method surpasses the state-of-the-art by a clear margin, on short video generation tasks, and can generate videos up to 40 frames. In contrast, the other methods collapse, which proves the effectiveness of our method in long-term video generation. 
We show qualitative results in Figure~\ref{fig:visual conparison of local region} and compare the video quality with previous methods on the same clip. Our method maintains consistent spatial and temporal appearance where the previous methods fail. 

\paragraph{Visualization of multi-view long videos generated by Delphi.}

\begin{figure}[t]
    \centering
    % \vspace{-0.5cm}
      % \includegraphics[width=1.0\linewidth]{figures/image_experiment/supp_ours_long_video.png}
      \includegraphics[width=1.0\linewidth]{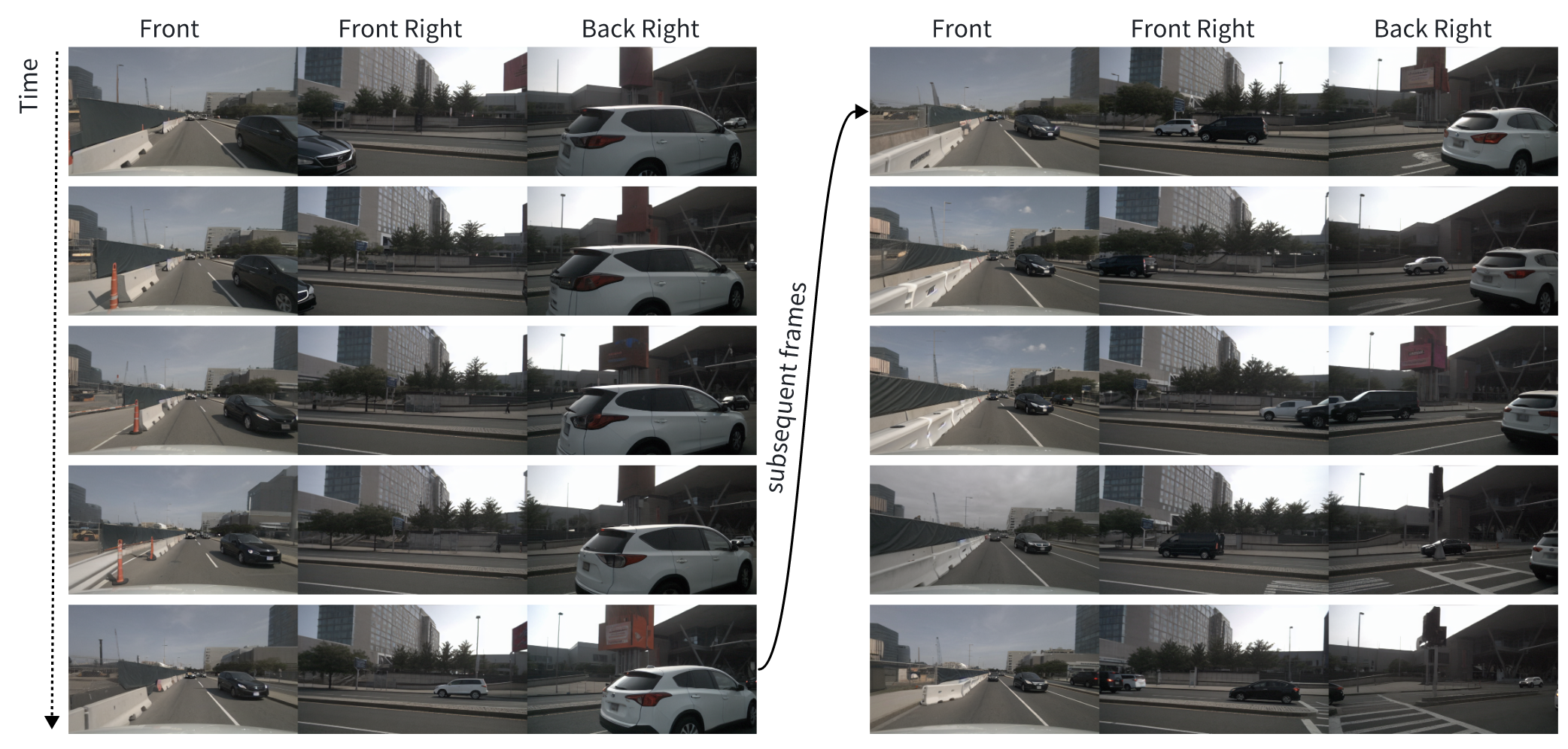}
    \vspace{-0.5cm}
    \caption{
    \textbf{The multi-view long video with spatiotemporal consistency generated by Delphi.}
    }
    % \vspace{-1em}
    \label{fig:supp:supp_ours_long_video} 
    \vspace{-0.3cm}
\end{figure}

\vspace{-0.1cm}
We demonstrate the generated multi-view long video in  Figure~\ref{fig:supp:supp_ours_long_video}. It can be seen that our method has the powerful ability to generate long videos with spatiotemporal consistency.

\vspace{-0.1cm}
\paragraph{Visualization comparison of multi-view video generated by different models.}
\begin{figure}[t]
    \centering
    \vspace{-0.5cm}
      \includegraphics[width=1.0\linewidth]{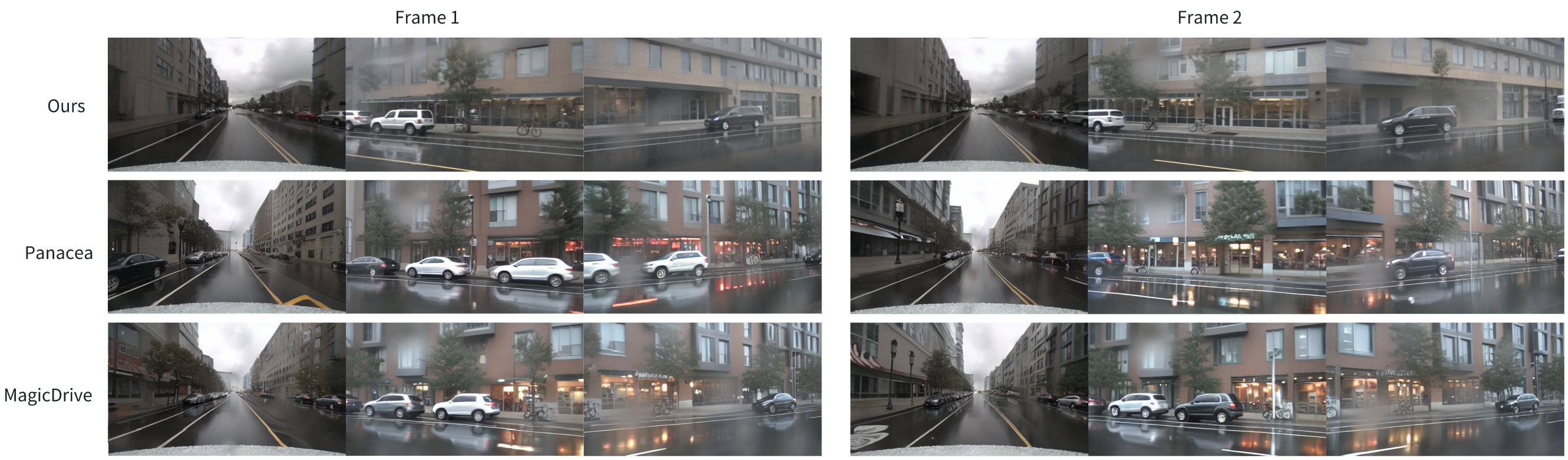}
    \vspace{-0.5cm}
    \caption{
    \textbf{Visualization comparison of multi-view video generated by different models.}
    }
    % \vspace{-1em}
    \label{fig:supp:supp_2frames_comparison} 
    \vspace{-0.2cm}
\end{figure}
We demonstrate visualization comparisons of multi-view video generated by different models in  Figure~\ref{fig:supp:supp_2frames_comparison}. It can be seen that our method has the powerful ability to generate long videos with spatiotemporal consistency.

% \subsection{Data Sampling Strategy of Failure-case Driven framework}
\subsection{Our failure-case driven framework boosts the end-to-end planning model}
% \paragraph{Random Sampling}
% \paragraph{Failure-case Driven Sampling}

% To compare the impact of different data sampling strategies, various numbers of data cases, different data engines, and different data sources on the performance of the end-to-end model within the failure-case driven framework, 

\begin{figure}[t]
    \centering
    \vspace{-0.2cm}
    \includegraphics[width=1.0\linewidth]{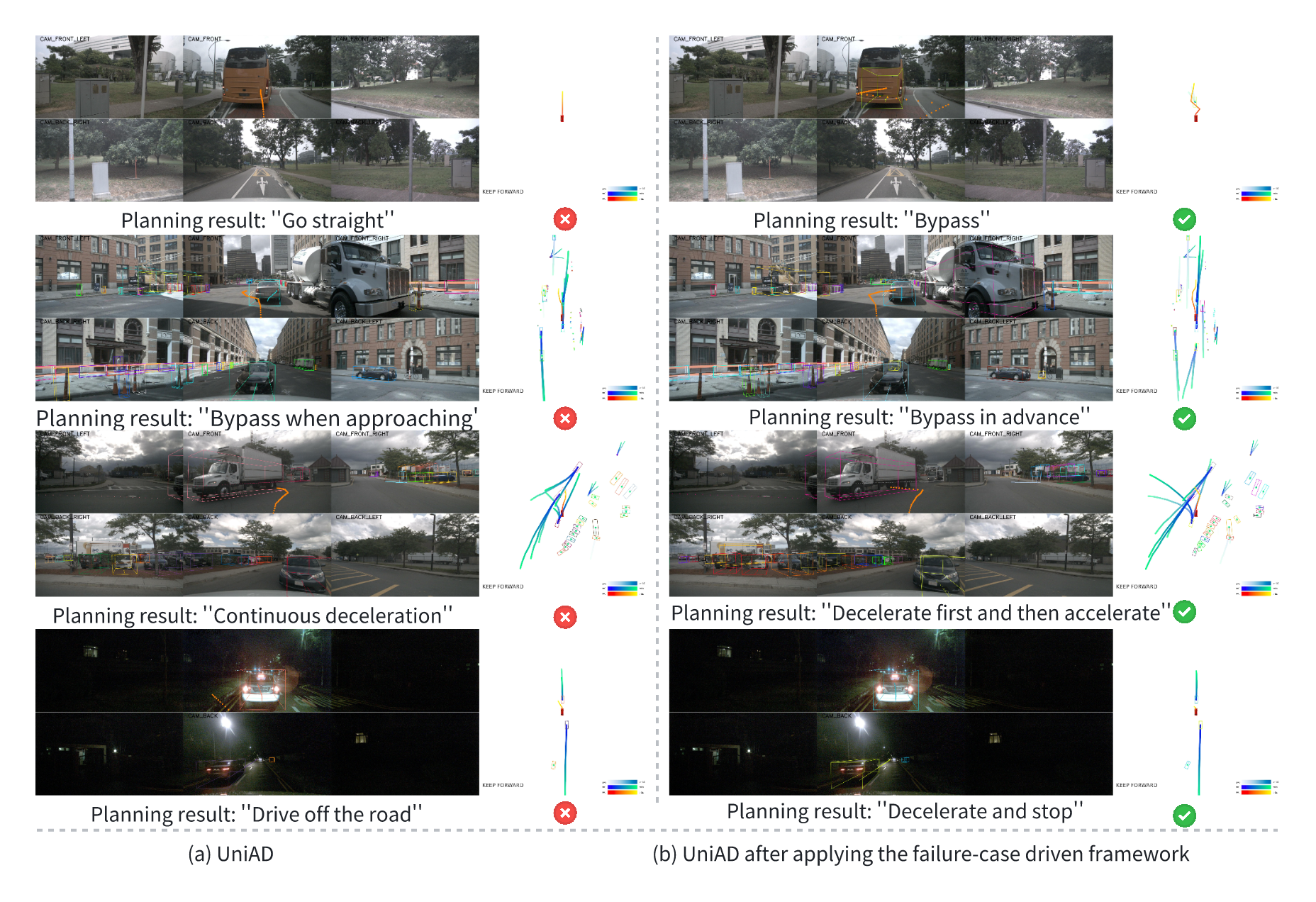}
    \vspace{-0.7cm}
    \caption{
    \textbf{Visualization of four examples before and after.} \textbf{(a)} Here, we show four hard examples from the validation set, ``large objects in the front'' and ``unprotected left turn at intersection''. \textbf{(b)} Our framework is able to fix these four examples without using these data during training. 
    }
    % \vspace{-1em}
    \label{fig:supp:uniad generalization comparison} 
    \vspace{-0.2cm}
\end{figure}

To prove the effectiveness of our framework, we compare three factors in Table~\ref{tab:failure_case_driven_framework}, the number of generated cases, data engine~(video generation method), and the choice of data source. In summary, we discover that, by generating only 4\% of the training set size data, our method can reduce the collision rate from 0.33 to 0.27 by a margin of 25\%. However, under the same setting, the collision rate increases if we use other data engines such as Panacea to fine-tune the UniAD. Nonetheless, we also exploit random sampling for both data engines and our method constantly outperforms the baseline. We also show how our frameworks can fix failure cases in Figure~\ref{fig:supp:uniad generalization comparison}.

% \paragraph{Generalization performance of end-to-end model with failure-case Driven Sampling.}
% \input{figures/supp_figures/uniad generalization comparison}

% Figure~\ref{fig:supp:uniad generalization comparison} shows the visual comparison of the planning results of the end-to-end model in some complex scenarios, including large objects nearby, occlusion scenes,  complex traffic  at cross intersection, and difficult light at night. We can see that our failure-case driven framework effectively enhances the model's generalization capability in these complex scenarios. 

% These experiments aimed to understand how these variables influence the overall effectiveness and performance of the end-to-end model. 

% In short, w
% Figure~\ref{fig:uniad_generalization_comparison} presents several extreme samples from the validation set. It is evident that the end-to-end model, trained using the failure-case driven framework, demonstrates significant improvements in planning performance in these extreme scenarios. This enhancement effectively increases the model's reliability and safety. 

% \input{tables/tex/failure_case_driven_framework}
\setlength{\tabcolsep}{11.0pt}
\renewcommand{\arraystretch}{1.2}

\begin{table*}[t!]
\centering
% \vspace{-0.1in}
\caption{
Performance comparison of the end-to-end models fine-tuned from the UniAD open source model by applying different data sampling strategies, numbers of data cases, data engines, and data sources in the failure-case driven framework.The baseline performance is presented in the first row of the table.
}

\scriptsize
\vspace{-0.1cm}
\resizebox{\textwidth}{!}{
\begin{tabular}{c|c|c|c|cccc}
\toprule
\multirow{2}{*}{\textbf{Sampling Strategy}}                                                                & \multirow{2}{*}{\textbf{Num of Cases}}  & \multirow{2}{*}{\textbf{Data Engine}} & \multirow{2}{*}{\textbf{Data Source}}  & \multicolumn{4}{c}{\textbf{Col. Rate(\%)$\downarrow$}} \\
                                                                                         &                                &                              &                               & \textbf{1s}     & \textbf{2s}     & \textbf{3s}    & \cellcolor{gray!15} \textbf{Avg.}    \\ \midrule

Baseline(UniAD) & -- & -- & -- & 0.10 & 0.18 & 0.71 &\cellcolor{gray!15} 0.33 \\ \hline  

\multirow{4}{*}{Random Sampling}                                                         & \multirow{2}{*}{14065 (50\%)}  & Panacea                      & \multirow{4}{*}{Training Set} &    \textbf{0.03}    &    0.23    &      0.79 &\cellcolor{gray!15} 0.35        \\
                                                                                         &                                & Delphi                       &                               &   0.08     &  0.20      & 0.58      & \cellcolor{gray!15} 0.29       \\
                                                                                         \cline{2-3} \cline{5-8}
                                                                                         & \multirow{2}{*}{28130 (100\%)} & Panacea                      &                               &  0.08      &  0.22      & 0.98      &\cellcolor{gray!15} 0.43        \\
                                                                                         &                                & Delphi                       &                               &  0.07      &  0.29      & 0.65      &\cellcolor{gray!15} 0.33        \\
                                                                                         \cline{1-8}
\multirow{3}{*}{\begin{tabular}[c]{@{}l@{}}Failure-case \\ Driven Sampling\end{tabular}} & \multirow{2}{*}{972 (4\%)}     & Panacea                      & \multirow{2}{*}{Training Set} &    0.05    &    0.18    &      0.81 &\cellcolor{gray!15}      0.35   \\
                                                                                         &                                & Delphi                       &                               &  0.08      &  0.18      &\textbf{0.56}       &\cellcolor{gray!15} \textbf{0.27}      \\
                                                                                         \cline{2-8}
                                                                                         & 429 (1.5\%)                              & Delphi                       & Validation Set                &  0.07      & \textbf{0.10}       &  0.61     &\cellcolor{gray!15} \textbf{0.26}     \\  
                                                                                         \bottomrule
\end{tabular}
}
\vspace{-0.3cm}
\label{tab:failure_case_driven_framework}
\end{table*}

\begin{figure}[t!]
    \centering
    \vspace{-0.1cm}
      \includegraphics[width=1.0\linewidth]{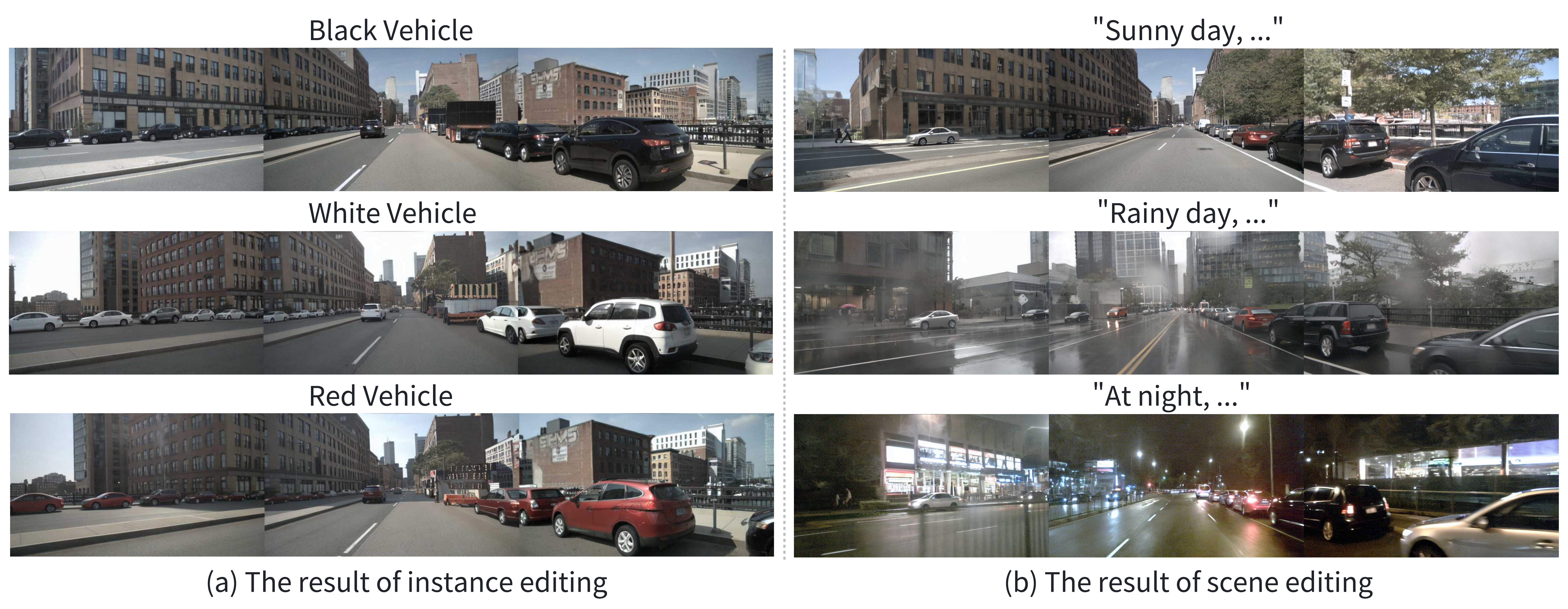}
    \vspace{-0.5cm}
    \caption{
    \textbf{Visualization of instance and scene editing. }
    % We demonstrate the precise controllability of Delphi.
    \textbf{(a)} shows the instance-level control result, such as the appearance attributes of all vehicles. \textbf{(b)} shows the scene-level control result, including weather and time.}
    \label{fig:visual precise control} 
    \vspace{-0.5cm}
\end{figure}

\mypara{What if we sample the layouts from the validation set?} 
Since our Delphi only sees the training set of nuScenes, a natural question is, can we include the validation set to see if we can further boost the performances of downstream tasks? Here, we collect failure cases from both the training and validation sets. Note that, since only layouts and captions are used in our framework, \emph{the validation video clips are never exposed in any training processes}. We notice that the collision rate is reduced from 0.33 to 0.26 with merely generating 429 cases, which is only 1.5\% of the training set size. This result might be interesting to industrial practitioners that a diffusion-based approach that only sees training dataset videos can effectively boost the performance of the validation set with layout and captions. 
% We used different sources of data, including the training set and the validation set. The last row of Table~\ref{tab:failure_case_driven_framework} presents a special experimental setup where the end-to-end model is directly trained on generated data from the failure cases in the validation set, which might be of interest to the industry. It is important to clarify that the end-to-end model was trained using the BEV layouts corresponding to the failure cases, and thus did not see the actual samples from the validation set during training. 
% This setup achieved the best performance, with a collision rate of 0.27$\%$. This further illustrates that feeding the model data related to failure cases can quickly improve efficiency.

\subsection{Ablation Studies}

We perform ablation studies to showcase the effectiveness of our method. 

% \input{tables/tex/sim2real_gap}
% sim2real_gap
\setlength{\tabcolsep}{7.0pt}
\renewcommand{\arraystretch}{1.2}
\begin{wraptable}{r}{6.0cm}
\centering
% \vspace{-0.27in}
\vspace{-0.5cm}
\caption{Performance of end-to-end model with real and generated data. We train the second stage of UniAD~\cite{hu2023uniad} 
with the officially released weights of the first stage as a starting point.
% In the first row, the results of StreamPETR and CVT are reported in Panacea~\cite{wen2023panacea} and MagicDrive~\cite{gao2023magicdrive}, respectively.
}
% \vspace{-0.1cm}
\scriptsize
\begin{tabular}{l|cccc}  
\toprule  
\multirow{2}{*}{\textbf{Method}} & \multicolumn{4}{c}{\textbf{Col. Rate(\%)}$\downarrow$}  \\  
& \textbf{1s} & \textbf{2s} & \textbf{3s} & \cellcolor{gray!15}\textbf{Avg.} \\  
\midrule  
Real & 0.07 & 0.24 & 0.70 & 0.34 \cellcolor{gray!15} \\    
Generated & 0.17 & 0.37 & 0.97 & 0.50 \cellcolor{gray!15} \\    
\hline  
Real$+$Real & \textbf{0.03} & 0.33 & 0.77 & 0.38 \cellcolor{gray!15} \\   
% \rowcolor{gray!15}   
Real$+$Generated & 0.08 & \textbf{0.18} & \textbf{0.56} & \textbf{0.27} \cellcolor{gray!15} \\    
\bottomrule  
\end{tabular}
% \vspace{-0.1in}
\label{tab:sim2real_gap}
\vspace{-0.7cm}
\end{wraptable}

\mypara{Ablating Sim-Real Gap.}
To further evaluate the sim-to-real gap, we train the UniAD with different portions of synthetic data. At the top of Table~\ref{tab:sim2real_gap}, we train UniAD with purely generated video clips, and the collision rate increases from 0.34 to 0.50. This indicates that the synthetic data cannot yet fully replace the real data. By contrast, if we consider the incremental learning setting, while we train UniAD with additional data, using synthetic data results in a much better performance while using additional real data deteriorates the performance from 0.34 to 0.38.
% In Table~\ref{tab:sim2real_gap}, we conduct a series of experiments to compare the gap between generated data and real data. Specifically, based on the layout annotations from the training set, we used Delphi to obtain a generated version of the training set. As shown in the second row of Table~\ref{tab:sim2real_gap}, the end-to-end model trained entirely on the generated training data achieved 68\% of the performance of the model trained on the real dataset. This effectively demonstrates a small gap between the generated data and the real data.
    % Furthermore, as shown in the fourth row, mixing the generated training data with the original training set resulted in a 20.6\% improvement in collision rate performance. However, increasing the amount of training data inevitably leads to longer training times. To eliminate the factor of training duration, we perform data augmentation using the real training data and train the model. As shown in the third row, even with the same extended training duration, data augmentation with real data does not improve performance and instead slightly degrades it.

% \input{tables/tex/generalization_enhancement_of_scene_editing}
\setlength{\tabcolsep}{4.4pt}
\renewcommand{\arraystretch}{1.2}
\begin{wraptable}{r}{5.2cm}
\centering
\vspace{-0.4cm}
\caption{
The effectiveness of Delphi's precise controllability on end-to-end models.
}
\vspace{-0.1cm}
\scriptsize
\begin{tabular}{cc|cccc}  
\toprule  
{\textbf{Scene}} &{\textbf{Instance}} & \multicolumn{4}{c}{\textbf{Col. Rate(\%)}$\downarrow$}  \\  
\textbf{editing} & \textbf{editing}& \textbf{1s} & \textbf{2s} & \textbf{3s} & \cellcolor{gray!15}\textbf{Avg.} \\  
\midrule
  &  &  0.10  &  0.20  &  0.71  &    0.34     \\  
 \checkmark & &0.11  &0.18   &0.64    &0.31      \\  
  &\checkmark  &  0.05  &  0.20  &  0.62  &   0.29   \\ \rowcolor{gray!15}  
\checkmark &\checkmark  &\textbf{0.08} & \textbf{0.18} & \textbf{0.56} & \textbf{0.27}      \\  \bottomrule
\end{tabular}
\vspace{-0.1in}
\label{tab:generalization_enhancement_of_scene_editing}
\vspace{-0.1in}
\end{wraptable}

\mypara{Ablating Scene and Instance Editing.} Table~\ref{tab:generalization_enhancement_of_scene_editing} demonstrates the effectiveness of data diversity for end-to-end models. Specifically, we edited existing scenes in two approaches: scene-level editing and instance-level editing. This fancy function allows us to generate a large amount of new data from a limited amount of existing data. As shown in Table~\ref{tab:generalization_enhancement_of_scene_editing}, simultaneous editing of both the scene and instances yields the best performance. 
Leveraging powerful precise controllability, Delphi maximizes end-to-end model performance by generating richer and more diverse data.

\setlength{\tabcolsep}{2.0pt}
\renewcommand{\arraystretch}{1.2}
\begin{wraptable}{r}{4.6cm}
    \centering
    \vspace{-0.5cm}
    \caption{Ablation study results for our proposed NRM and FTCM.}
    \vspace{-0.2cm}
    \scriptsize
        \begin{tabular}{l|ccc}
            \toprule
            Method & ${\text{FID}}\downarrow$ & ${\text{FVD}}\downarrow$ & $ {\text{CLIP}} \uparrow $ \\ \rowcolor{gray!15}  \midrule
            Delphi  &\textbf{15.08}   &\textbf{275.6}   &\textbf{86.73} \\ \hline
            w/o NRM &19.81  &291.5  &85.22  \\
            w/o NRM $\&$ FTCM &22.85  &346.96 &82.91   \\
            \bottomrule
        \end{tabular}
    \label{tab:ablation}
    \vspace{-0.3cm}
\end{wraptable}

\mypara{Ablating NRM and FTCM.} 
% We conducted an ablation study on the two key modules in Delphi: the Noise Reinitialization Module (NRM) and the Feature-aligned Temporal Consistency Module (FTCM) in Table~\ref{tab:ablation} .
In Table~\ref{tab:ablation}, we validate the two modules, the Noise Reinitialization Module (NRM) and the Feature-aligned Temporal Consistency Module (FTCM). We see an evident increase in all metrics to validate the effectiveness of our proposed method. In particular, the FTCM  structure improves FID from 22.85 to 19.81 while the NRM further boosts it.

\section{Conclusion}

In summary, we propose a novel video generation method for autonomous driving scenarios that can synthesize up to 40 frames of videos on nuScenes dataset. Surprisingly, we show that with a diffusion model trained only with training split, we are able to improve the performance of the end-to-end planning model by a sample efficient failure-case driven framework. We hope to shed light on addressing the data scarcity problem for both researchers and practitioners in this field, and make a solid step towards making autonomous driving vehicles safe on the road.

\mypara{Limitation and Societal Impact.}
Our \emph{Delphi} takes BEV layout as input to ensure the control ability, i.e. we are only capable of enriching the appearances and cannot change the layout during the synthesis processes. This leads to a limitation that our framework can be only used in an open-loop setting{\cite{caesar2020_nuscenes} but not the close-loop one. Yet, another limitation is when the end-to-end model performs perfectly in the training dataset, our failure-case driven sampling does not work. In terms of societal impact, we believe our method can be used to boost the performance of end-to-end models and may help the deployment of large-scale autonomous driving vehicles in the future.

\bibliographystyle{plainnat}
\bibliography{11_references}

\newpage
\appendix

% \appendix
% \section{Supplemental material}

\section{Method}
\subsection{Detailed implementation of failure case driven framework}

\paragraph{Collecting Failure Cases.}
Initially, we assess the performance of the base end-to-end model on the validation set . For this evaluation, we utilize the UniAD~\cite{hu2023uniad} base model as our starting point. We employ a metric, wherein a collision occurring within 3 seconds on the path planned by the algorithm qualifies a scenario as a failure case. Additionally, to gain further insights, we visualize both the perception results of the 3D boxes and the planning outcomes derived from the end-to-end model. Through this process, we identify and select failure cases for further analysis.

\paragraph{Analyzing data pattern.}
We initially anticipated that large visual-language models would be able to automatically pinpoint the reasons behind algorithm failures. However, our investigations revealed that a straightforward inquiry was insufficient for this purpose. Consequently, we devised a multi-round inquiry method leveraging VLM~\cite{achiam2023gpt}. This approach enables a more precise analysis of the factors contributing to algorithm failures, as well as a detailed description of the key elements leading to such failures.

Specifically, we feed the visualization outcomes from the preceding step into VLM and prompt it to discern whether the primary cause of failure stems from perception or planning issues. In cases where perception is the culprit, the reasons can be further categorized into various factors such as nighttime darkness, challenges in recognizing large nearby objects, or the inability to identify rare object categories. On the other hand, if planning is identified as the source of failure, VLM can differentiate between scenarios like occlusion or infrequent interactions, including running a red light or crossing the road. Ultimately, based on the previously established reasons for failure, we prompt VLM to offer a precise account of the specific factors that led to the failure.

\paragraph{Retrieving similar scenes.}
Using the detailed image description, we employ BLIP-2~\cite{LI2023blip2} to identify and retrieve scenes from the train set that closely correspond to the reasons behind the failure. This process involves quantifying the cosine similarity between embeddings extracted from both the image and the designated text input using BLIP-2. Based on this similarity measure, we then select and retrieve only the top-$k$ most relevant images.

\paragraph{Updating Model.}
Based on the identified potential failure scenarios, we created an extensive and varied image dataset utilizing \emph{Delphi}. We obtained scene captions from VLM using corresponding sample tokens and employed these captions as input to generate analogous images with \emph{Delphi}. 

To augment data diversity, we employed a LLM to adjust the captions inputted into \emph{Delphi}. This approach facilitated the alteration of scene descriptions to encompass various scene conditions and instance conditions, such as sunny, rainy, cloudy, Night, suburban, changing the color of the cars. Consequently, feeding these revised captions into \emph{Delphi} resulted in the generation of a broader range of images.

However, we discovered that directly utilizing the generated failure scenes for training could result in overfitting. While the trained model excelled in the selected failure instances, its performance suffered in previously successful cases. Therefore, to mitigate this issue, we integrated our generated data with the complete train set for each fine-tuning session. This strategy proved effective in optimizing the model's overall performance.

Ultimately, we trained the end-to-end model using this combined dataset, yielding a refined model that marked the commencement of the subsequent iteration of the improvement cycle.

\section{Experiments}

\subsection{Metrics}

\paragraph{Metrics about Quality and Controllability of Generated Video.}
We evaluate the quality of the generated videos from two aspects: quality and controllability. 
Specifically, for quality, we use Frechet Inception Distance (${\text{FID}}$)~\cite{heusel2017_fid} to assess the realism of single-frame single-view images in the generated videos, Frechet Video Distance (${\text{FVD}}$)~\cite{unterthiner2018fvd} to evaluate the temporal consistency of single-view videos, and CLIP scores (${\text{CLIP}}$)~\cite{yang2023bevcontrol} to assess the spatial consistency of single-frame multi-view images. 
For controllability, we utilize the popular BEV detection model StreamPETR~\cite{wang2023streampetr} and end-to-end model~\cite{hu2023uniad} to evaluate the generated data and report the NDS score and the Average Collision Rate(Avg. Col. Rate) respectively, which comprehensively reflects the geometric alignment between the generated images and the BEV layout annotations. By using these evaluation metrics, we can ensure that the generated results maintain high standards in both quality and controllability.

 % We use NDS Score to access the controllability of \emph{Delphi}, which represents the geometric alignment between the generated multi-view videos and BEV sketch sequences.
 % \vspace{-0.5em} 
\paragraph{Metrics about Effectiveness of the Generated Video for End-to-End Model.}
To evaluate the effectiveness of our proposed failure-case driven framework based upon the Delphi for the end-to-end model, we utilize the generated diverse training data to augment the end-to-end model's origin training data. 
Specifically, we evaluate the performance of the end-to-end model by applying data augmentation on the nuScenes validation set and report the average collision rate.

\subsection{More Experimental Details}
\paragraph{Experimental Setting of the end-to-end model.}
During the training phase, we utilize the model available on the UniAD official repository as our foundation for fine-tuning. To enhance the training process, we have decreased the learning rate by a factor of 10, setting it to 2e-5. Additionally, we maintain consistency with the hyperparameters recommended on the UniAD repository, including the optimizer settings.

\paragraph{Computation Efficiency and Hardware Requirements~~}
We report the model complexity of our two model variants in Table~\ref{tab:supp_model_efficiency}. 
% Since the main computation lies in adding noise and denoising of diffusion model, our Controller and Coordinator are lightweight, accounting only account for 20\% and 24\% of the overall parameters and generation time.
We will further provide the generated data on the nuScenes training set for the convenience of data augmentation.

\setlength{\tabcolsep}{6pt}
\renewcommand{\arraystretch}{1.2}
  \begin{table}[h!]
% \vspace{-0.1in}
  \caption{Model efficiency and hardware requirements.}
  \label{tab:supp_model_efficiency}
% \vspace{-0.05in}
    \centering
    \scriptsize
    \begin{tabular}{c|c|c|c|c}
    % \hline
    \toprule
 \textbf{Model}& \textbf{Parameter} & \textbf{Inference Memeory\&GPU} & \textbf{Inference Time} & \textbf{Train config} \\ 
 % \hline 
 \midrule
 multi-view single-frame & 0.5B  &22GB(RTX3090)  & 4s / example & 8$\times$A100, 24 hours        \\ 
 
 multi-view multi-frame & 1.1B  &39GB(A100 40G)  & 4s / example & 8$\times$A800, 72 hours        \\ 
 % \hline
 \bottomrule
    \end{tabular}
    \vspace{-0.1in}
\end{table}

% \subsection{More Visualization Results}

\subsection{Validating each components of our failure-case driven framework}

As in Table~\ref{tab:failure_case_driven_framework}, we compare in three aspects, data sampling strategy, number of generating cases and data engine validation. 

\paragraph{Data Sampling Strategy.} We evaluated different data sampling strategies, such as random sampling and failure-case targeted sampling. In the upper part of Table~\ref{tab:failure_case_driven_framework}, we randomly selected various proportions of data samples from the training dataset and used the corresponding BEV layout and original scene captions to generate new data. In the lower part of Table~\ref{tab:failure_case_driven_framework}, we retrieved training data with similar patterns to failure cases from the validation set and generated diverse weather data using the powerful control capabilities of the generative model. The newly generated data was mixed with the original data to train the end-to-end model. It was observed that the end-to-end model, enhanced through failure-case guided data augmentation, achieved the best performance. This demonstrates that the end-to-end model is under-trained in these failure cases, and feeding it more failure-case related training data can achieve optimal generalization performance with fewer computational resources. 

\paragraph{Numbers of Cases.} We investigated the quantity of data samples. We randomly sampled 14,065 and 28,130 training samples (approximately 50$\%$ and 100$\%$ of the entire training set) from the training set. The results generated by the configuration of the generative model on these samples were used for data augmentation. As shown in the upper part of Table~\ref{tab:failure_case_driven_framework}, the performance of the end-to-end model worsened as the number of samples increased. This indicates that using training data with a style similar to the original training set can only help the model to a limited extent. Thus, it prompted us to consider increasing the diversity of the training data.

\paragraph{Data Engine.}  
We tested various data generation engines, including Delphi and other state-of-the-art generative models Panacea~\cite{wen2023panacea}, to compare their effectiveness in generating high-quality training data for model enhancement. From the three sets of comparison experiments, it can be seen that the data generated by Delphi effectively improves the performance of the end-to-end model compared to other generative models. This is due to Delphi's superior fine control capabilities in scene generation, leading to more diverse training data for model tuning.

\ifbool{isArxiv}{}{
\newpage
\section*{NeurIPS Paper Checklist}

\begin{enumerate}

\item {\bf Claims}
    \item[] Question: Do the main claims made in the abstract and introduction accurately reflect the paper's contributions and scope?
    \item[] Answer: \answerYes{} % Replace by \answerYes{}, \answerNo{}, or \answerNA{}.
    \item[] Justification: We made main claims in the abstract and introduction to accurately reflect the paper's contributions and scope.

\item {\bf Limitations}
    \item[] Question: Does the paper discuss the limitations of the work performed by the authors?
    \item[] Answer: \answerYes{} % Replace by \answerYes{}, \answerNo{}, or \answerNA{}.
    \item[] Justification: We discussed the limitations of the work in the conclusion section.

\item {\bf Theory Assumptions and Proofs}
    \item[] Question: For each theoretical result, does the paper provide the full set of assumptions and a complete (and correct) proof?
    \item[] Answer: \answerNA{} % Replace by \answerYes{}, \answerNo{}, or \answerNA{}.
    \item[] Justification: \ky{We do not have theoretical results in our paper.}
    % We provided the full set of assumptions and a complete (and correct) proof for each theoretical result.

    \item {\bf Experimental Result Reproducibility}
    \item[] Question: Does the paper fully disclose all the information needed to reproduce the main experimental results of the paper to the extent that it affects the main claims and/or conclusions of the paper (regardless of whether the code and data are provided or not)?
    \item[] Answer: \answerYes{} % Replace by \answerYes{}, \answerNo{}, or \answerNA{}.
    \item[] Justification: We fully disclosed all the information needed to reproduce the main experimental results of the paper and supplementary materials.

\item {\bf Open access to data and code}
    \item[] Question: Does the paper provide open access to the data and code, with sufficient instructions to faithfully reproduce the main experimental results, as described in supplemental material?
    \item[] Answer: \answerYes{} % Replace by \answerYes{}, \answerNo{}, or \answerNA{}.
    \item[] Justification: We will release code and data upon paper acceptance.

\item {\bf Experimental Setting/Details}
    \item[] Question: Does the paper specify all the training and test details (e.g., data splits, hyperparameters, how they were chosen, type of optimizer, etc.) necessary to understand the results?
    \item[] Answer: \answerYes{} % Replace by \answerYes{}, \answerNo{}, or \answerNA{}.
    \item[] Justification: We specified all the training and test details in the paper, code and supplementary materials.

\item {\bf Experiment Statistical Significance}
    \item[] Question: Does the paper report error bars suitably and correctly defined or other appropriate information about the statistical significance of the experiments?
    \item[] Answer: \answerYes{} % Replace by \answerYes{}, \answerNo{}, or \answerNA{}.
    \item[] Justification: We reported error bars about the statistical significance of the experiments.

\item {\bf Experiments Compute Resources}
    \item[] Question: For each experiment, does the paper provide sufficient information on the computer resources (type of compute workers, memory, time of execution) needed to reproduce the experiments?
    \item[] Answer: \answerYes{} % Replace by \answerYes{}, \answerNo{}, or \answerNA{}.
    \item[] Justification: For each experiment, we provided sufficient information on the computer resources needed to reproduce the experiments.
    
\item {\bf Code Of Ethics}
    \item[] Question: Does the research conducted in the paper conform, in every respect, with the NeurIPS Code of Ethics \url{https://neurips.cc/public/EthicsGuidelines}?
    \item[] Answer: \answerYes{} % Replace by \answerYes{}, \answerNo{}, or \answerNA{}.
    \item[] Justification: The research conducted in the paper conform, in every respect, with the NeurIPS Code of Ethics.

\item {\bf Broader Impacts}
    \item[] Question: Does the paper discuss both potential positive societal impacts and negative societal impacts of the work performed?
    \item[] Answer: \answerYes{} % Replace by \answerYes{}, \answerNo{}, or \answerNA{}.
    \item[] Justification: We discussed both potential positive societal impacts and negative societal impacts of the work performed.
    
\item {\bf Safeguards}
    \item[] Question: Does the paper describe safeguards that have been put in place for responsible release of data or models that have a high risk for misuse (e.g., pretrained language models, image generators, or scraped datasets)?
    \item[] Answer: \answerYes{} % Replace by \answerYes{}, \answerNo{}, or \answerNA{}.
    % \item[] Justification: \ky{We do not introduce new data in this paper, so we do not see any potential risk of misuse. }
    We described safeguards that have been put in place for responsible release of data or models that have a high risk for misuse. \ky{However, this will only be the case that paper is accepted when we release our code and model. }

\item {\bf Licenses for existing assets}
    \item[] Question: Are the creators or original owners of assets (e.g., code, data, models), used in the paper, properly credited and are the license and terms of use explicitly mentioned and properly respected?
    \item[] Answer: \answerYes{} % Replace by \answerYes{}, \answerNo{}, or \answerNA{}.
    \item[] Justification: We credited the creators or original owners of assets used in the paper.

\item {\bf New Assets}
    \item[] Question: Are new assets introduced in the paper well documented and is the documentation provided alongside the assets?
    \item[] Answer: \answerNA{} % Replace by \answerYes{}, \answerNo{}, or \answerNA{}.
    \item[] Justification: \ky{We do not introduce new assets.}
    % We provided documentation alongside the new assets introduced in the paper.

\item {\bf Crowdsourcing and Research with Human Subjects}
    \item[] Question: For crowdsourcing experiments and research with human subjects, does the paper include the full text of instructions given to participants and screenshots, if applicable, as well as details about compensation (if any)? 
    \item[] Answer: \answerNA{} % Replace by \answerYes{}, \answerNo{}, or \answerNA{}.
    \item[] Justification: \ky{We do not involve crowdsourcing nor research with human subjects.}
    % For crowdsourcing experiments and research with human subjects, we provided  the paper include the full text of instructions to participants and screenshots.

\item {\bf Institutional Review Board (IRB) Approvals or Equivalent for Research with Human Subjects}
    \item[] Question: Does the paper describe potential risks incurred by study participants, whether such risks were disclosed to the subjects, and whether Institutional Review Board (IRB) approvals (or an equivalent approval/review based on the requirements of your country or institution) were obtained?
    \item[] Answer: \answerNA{} % Replace by \answerYes{}, \answerNo{}, or \answerNA{}.
    \item[] Justification: \ky{The answer NA means that the paper does not involve crowdsourcing nor research with human subjects.}
    % We  described potential risks incurred by study participants and  disclosed to the subjects.
\end{enumerate}
}

\end{document}